\documentclass[runningheads]{llncs}
\usepackage[T1]{fontenc}
\usepackage{amsmath}
\usepackage{amsfonts}
\usepackage{dsfont}
\usepackage{graphicx}
\usepackage{tikz}
\usepackage{multirow}
\usepackage{array}
\usepackage{tabularx,booktabs}
\usepackage{subcaption}
\usepackage{dirtytalk}
\usepackage{bm}
\usepackage{dsfont}
\usepackage{float}
\usetikzlibrary{arrows.meta}
\usepackage{color}
\usepackage{hyperref}
\usepackage{rotating}
\usepackage[misc]{ifsym}

\begin{document}
\newcommand{\repeatthanks}{\textsuperscript{\thefootnote}}

\title{Towards Affordable Tumor Segmentation and Visualization for 3D Breast MRI Using SAM2}

\titlerunning{Affordable Tumor Segmentation and Visualization for 3D Breast MRI}

\author{
Solha Kang\inst{1,2}\thanks{Equal contribution.} \and
Eugene Kim\inst{1}\repeatthanks \and
Joris Vankerschaver\inst{1,2} \and\\
Utku Ozbulak\inst{1,3}(\Letter)
}

\authorrunning{Kang et al.}

\institute{
Center for Biosystems and Biotech Data Science, Ghent University Global Campus, Incheon, Republic of Korea
\and
Department of Mathematics, Computer Science and Statistics, Ghent University, Ghent, Belgium
\and
IDLab, Department of Electronics and Information Systems, Ghent University, Ghent, Belgium\\
(\Letter) \email{utku.ozbulak@ghent.ac.kr}
}

\maketitle
\begin{abstract}\let\thefootnote\relax\footnotetext{Accepted for publication in the 28th International Conference on Medical Image Computing and Computer Assisted Intervention (MICCAI), 2nd Deep Breast Workshop on AI and Imaging for Diagnostic and Treatment Challenges in Breast Care (DeepBreath), 2025.}
Breast MRI provides high-resolution volumetric imaging critical for tumor assessment and treatment planning, yet manual interpretation of 3D scans remains labor-intensive and subjective. While AI-powered tools hold promise for accelerating medical image analysis, adoption of commercial medical AI products remains limited in low- and middle-income countries due to high license costs, proprietary software, and infrastructure demands. In this work, we investigate whether the Segment Anything Model 2 (SAM2) can be adapted for low-cost, minimal-input 3D tumor segmentation in breast MRI. Using a single bounding box annotation on one slice, we propagate segmentation predictions across the 3D volume using three different slice-wise tracking strategies: top-to-bottom, bottom-to-top, and center-outward. We evaluate these strategies across a large cohort of patients and find that center-outward propagation yields the most consistent and accurate segmentations. Despite being a zero-shot model not trained for volumetric medical data, SAM2 achieves strong segmentation performance under minimal supervision. We further analyze how segmentation performance relates to tumor size, location, and shape, identifying key failure modes. Our results suggest that general-purpose foundation models such as SAM2 can support 3D medical image analysis with minimal supervision, offering an accessible and affordable alternative for resource-constrained settings.
\end{abstract}


\begin{figure}[t!]
\centering
\begin{subfigure}{0.225\textwidth}
\centering
\includegraphics[width=0.95\textwidth]{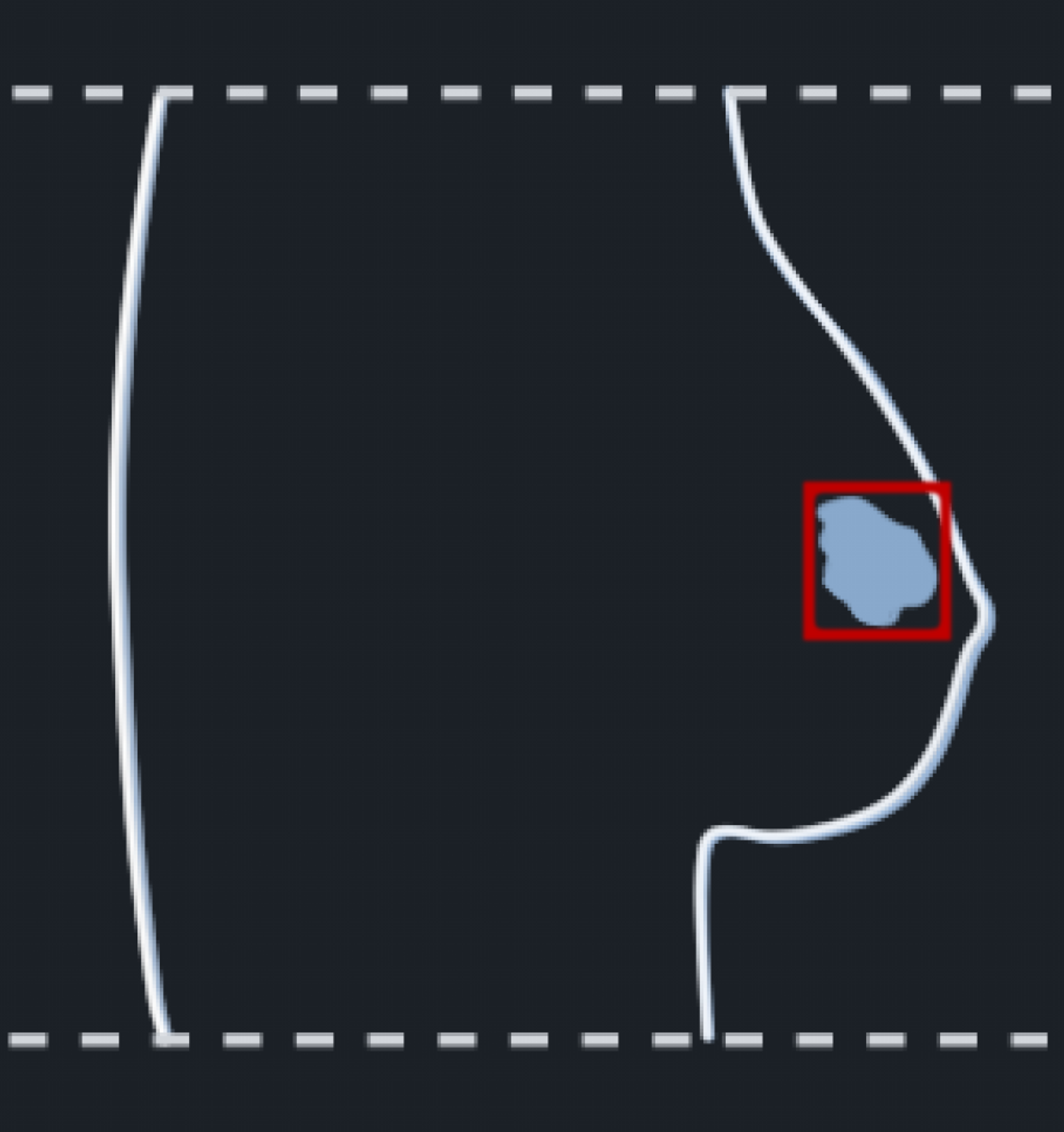}
\caption{Side profile}
\label{fig:MRI_chest_drawing}
\end{subfigure}
\begin{subfigure}{0.715\textwidth}
\centering
\includegraphics[width=0.158\textwidth]{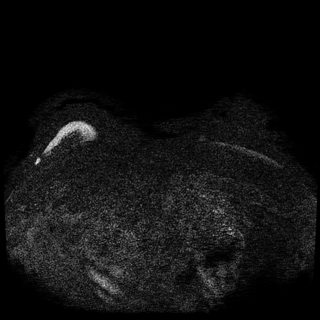}
\includegraphics[width=0.157\textwidth]{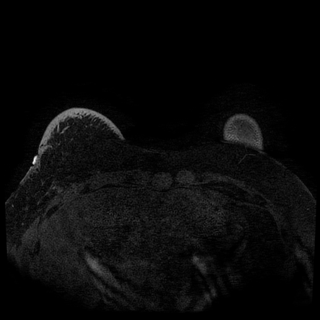}
\includegraphics[width=0.157\textwidth]{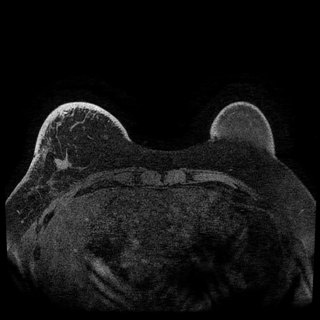}
\includegraphics[width=0.157\textwidth]{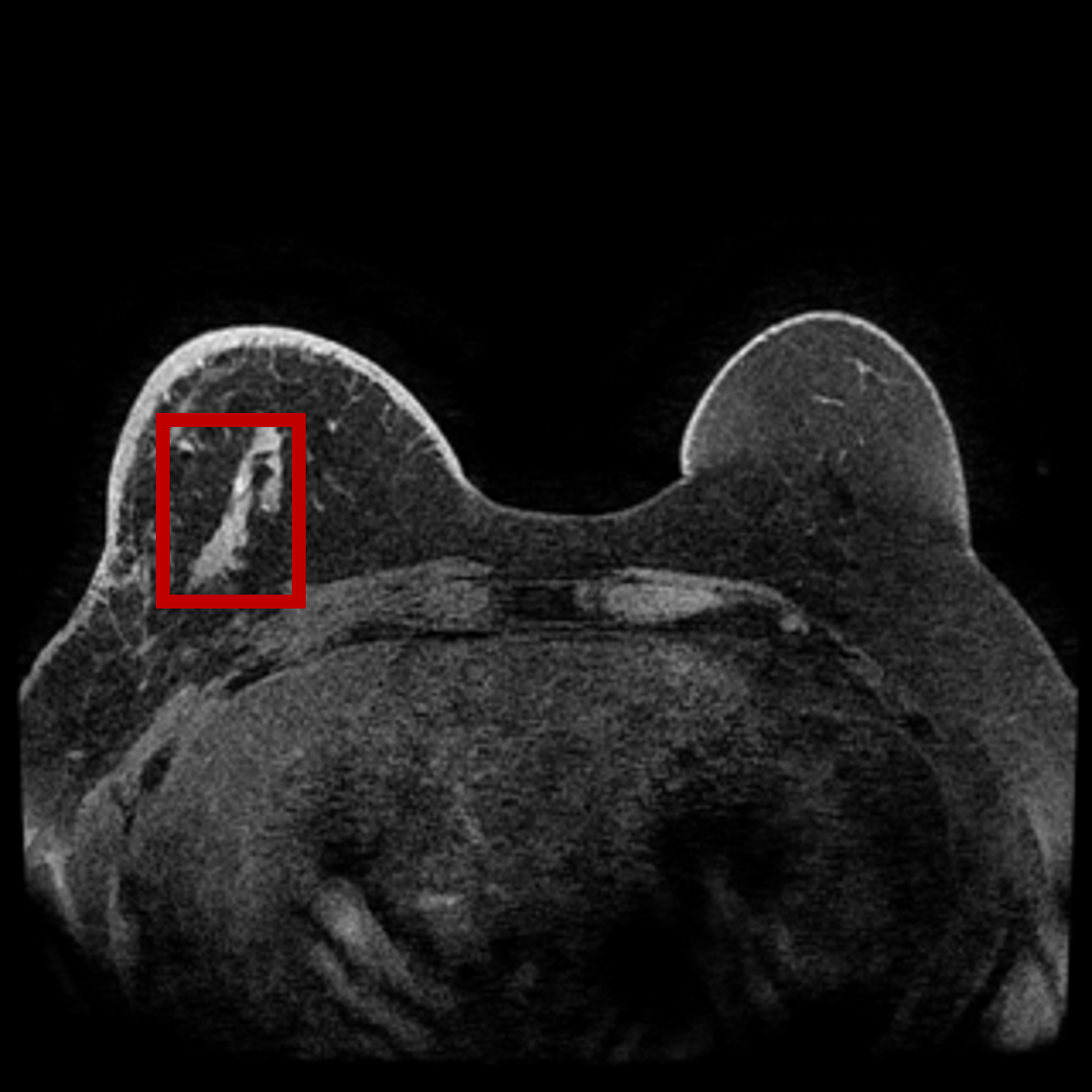}
\includegraphics[width=0.157\textwidth]{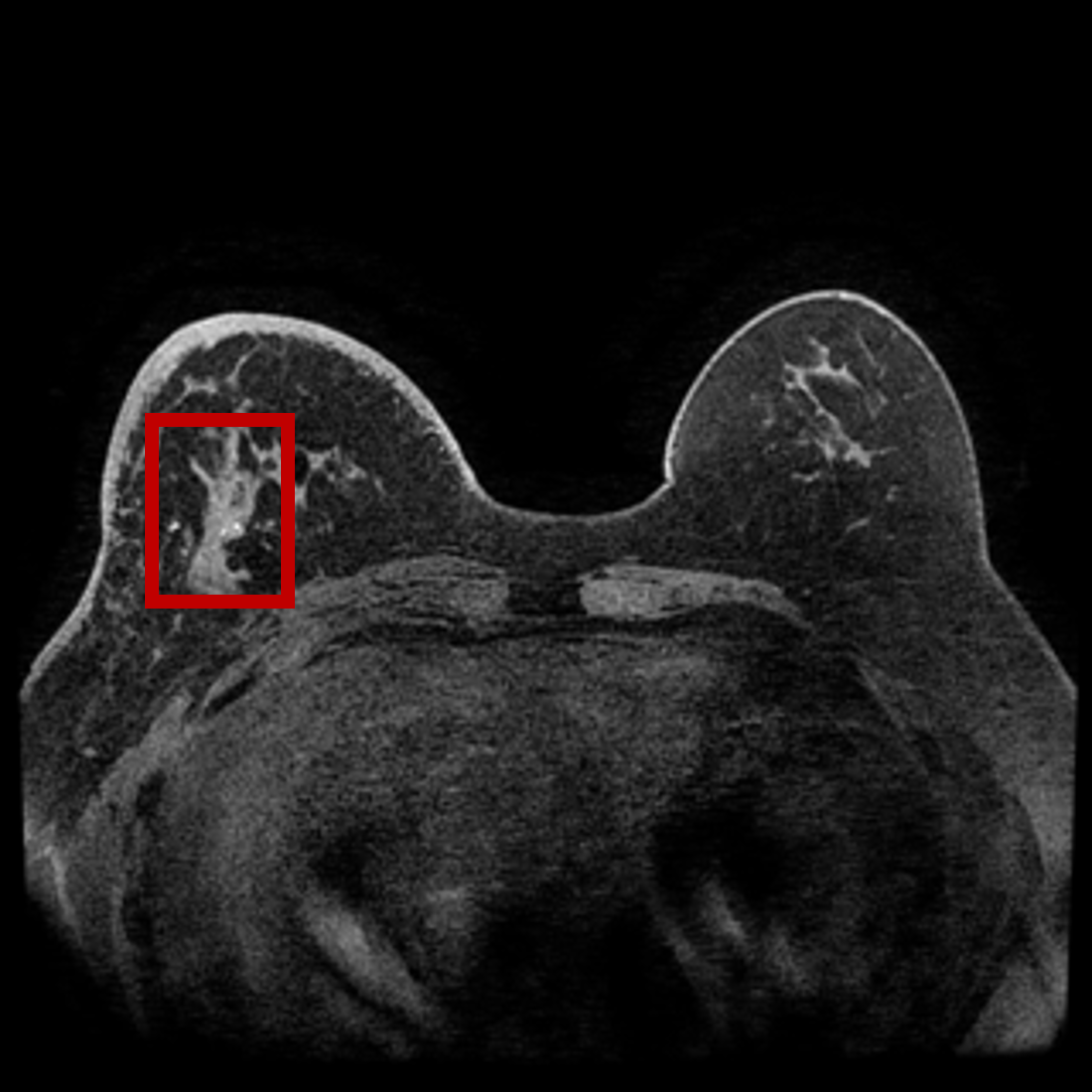}
\includegraphics[width=0.157\textwidth]{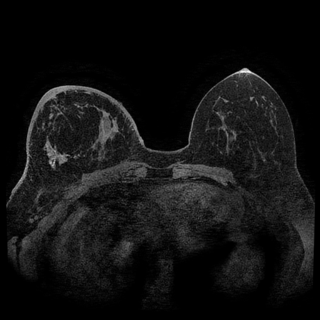}
\includegraphics[width=0.157\textwidth]{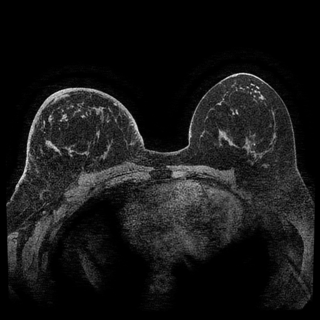}
\includegraphics[width=0.157\textwidth]{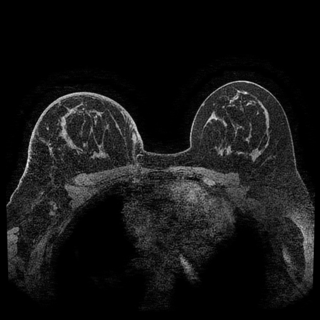}
\includegraphics[width=0.157\textwidth]{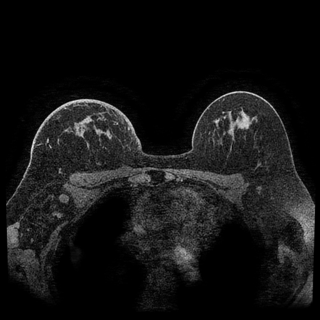}
\includegraphics[width=0.157\textwidth]{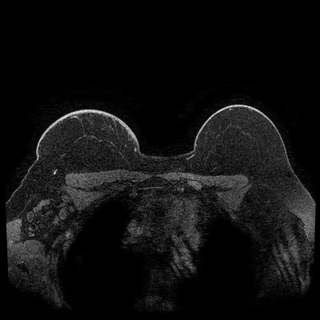}
\includegraphics[width=0.157\textwidth]{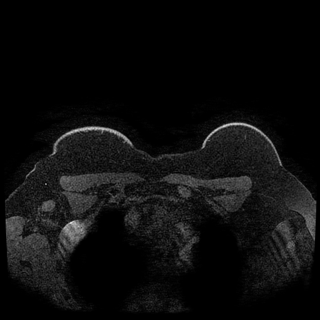}
\includegraphics[width=0.157\textwidth]{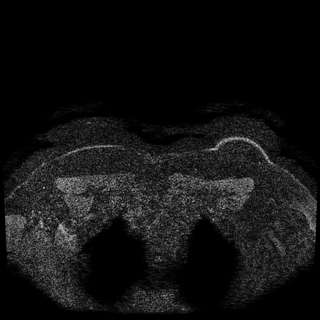}
\caption{Example MRI slices}
\label{fig:data_top}
\end{subfigure}
\caption{(a) A side profile diagram of the breast, highlighting the imaging region bounded by two dashed lines. The tumor is illustrated as a blue mass, with a red box indicating its location. (b) Example MRI slices obtained from a cross-sectional region of the breast, shown in sequential order. The slice positions correspond to the imaging region illustrated in (a), and the red box highlights the tumor for slices intersecting the 3D bounding box annotation.}
\label{fig:data_overview}
\end{figure}

\newpage
\section{Introduction}

Breast cancer is the most commonly diagnosed cancer worldwide and is a leading cause of cancer-related mortality, particularly among women~\cite{lukasiewicz2021breast,siegel2024cancer}. Early and accurate detection is critical for improving patient outcomes, and magnetic resonance imaging (MRI) plays a vital role in this process. In clinical settings, 3D breast MRIs provide high-resolution, volumetric information that can reveal subtle morphological features of tumors~\cite{clift2022current}. This makes them particularly useful in assessing dense breast tissue, evaluating tumor extent, and planning treatment. However, these volumetric scans often consist of hundreds of axial slices (see \figurename~\ref{fig:data_overview}), requiring clinicians to mentally reconstruct and interpret the full 3D context, a task that is time-consuming and prone to subjectivity~\cite{mango2015abbreviated,mootz2019changing}.

At the same time, the field of computer vision and artificial intelligence has seen rapid advancements in recent years, with powerful models and pre-trained architectures becoming increasingly available through open-source platforms~\cite{awais2023foundational,han2021pre,zhou2024comprehensive}. Techniques that once required significant technical resources are now widely accessible to researchers and developers. Despite this progress, the integration of AI into real-world healthcare systems, particularly in under-resourced settings, remains limited~\cite{alowais2023revolutionizing,kelly2019key}. Proprietary software, licensing fees, and infrastructure requirements often create a financial barrier, rendering AI-powered diagnostic tools inaccessible to many hospitals in low- and middle-income countries~\cite{dangi2025transforming,panch2019inconvenient}.

On the side of research, recent approaches to 3D breast tumor segmentation have primarily relied on supervised deep learning methods, including 3D convolutional neural networks (CNNs), U-Net variants~\cite{isensee2021nnu,zhang2024unimrisegnet}, and transformer-based architectures~\cite{chen2021transunet,hatamizadeh2022unetr}. These models typically require large annotated datasets and intensive training efforts, which limit their employability in many scenarios. While these methods have shown strong performance on benchmark datasets, their dependence on dense voxel-wise annotations and task-specific tuning presents a barrier to generalization and scalability, particularly in low-data or low-resource scenarios.

In this work, we examine how the recent advances in video object tracking~\cite{cheng2022xmem} can be harnessed to develop accessible, low-cost AI solutions for 3D breast MRI interpretation,  with a focus on tools that require minimal infrastructure (i.e., no training or fine-tuning). Beyond technical performance, our work emphasizes the importance of equitable access in medical AI. Building entirely on open-source tools and demonstrating their viability for 3D breast MRI segmentation and visualization, we advocate for affordable AI solutions that can be adopted across diverse healthcare systems.

To concretely realize this vision of accessible AI and following the works of~\cite{ma2025medsam2,zhu2024medical} we leverage recent advances in vision foundation models, specifically employing the open-source SAM2 model~\cite{ravi2024sam}, to segment and visualize breast tumors in 3D MRI volumes using only minimal human-in-the-loop input (a single bounding box). We show that, although not originally designed for volumetric data, SAM2 can effectively segment breast tumors across slices and generate clinically meaningful visualizations. At the same time, we identify and analyze failure modes in this setting, offering a critical examination of when and why the model breaks down. Our goal is to demonstrate that such tools can deliver practical and accurate results without relying on expensive commercial software or high-end infrastructure, making them viable for deployment in resource-constrained healthcare environments.

\section{Methodology}
\label{sec:methodology}

\subsection{Dataset}

The Duke Breast Cancer Dataset is a large-scale collection of pre-operative 3D breast MRI scans curated from 922 patients treated at Duke University Medical Center~\cite{duke_dataset}. The dataset includes 3D bounding box annotations of tumor locations for each cancer patient and is frequently used as a benchmark for detecting tumors in 2D MRI slices. For our task, we follow the procedure employed in prior work to extract 2D horizontal slice images from the pre-contrast DCE-MRI volumes
~\cite{chung2024evaluating,kang2024exploring,duke_intrinsic,duke_dataset_property}. This process results in over $140,000$ 2D images, averaging approximately $130$ slices per patient. Among these, tumor-positive slices occur in about $27$ images per patient, totaling more than $20,000$ images across the dataset.

In 2025, an expanded version of the dataset, referred to as the MAMA-MIA dataset, was released, featuring voxel-level tumor segmentations verified by expert reviewers~\cite{garrucho2025large}. This version includes approximately $9,000$ segmentation masks corresponding to tumor-positive MRI slices, averaging about $34$ masks per patient. In our study, we use the labeled images from MAMA-MIA dataset for $279$ patients.

\subsection{Model}

SAM2 is an open-source foundation model built on vision transformers for the purpose of zero-shot image and video segmentation~\cite{ravi2024sam}. It accepts prompts such as points, bounding boxes, or masks, and produces segmentation masks without requiring task-specific training. In this work, we leverage its tracking functionality to propagate a single-slice bounding box prompt across volumetric breast MRI data for 3D tumor segmentation.

\begin{figure}[t!]
\centering
\begin{subfigure}{0.225\textwidth}
\centering
\includegraphics[width=0.95\textwidth]{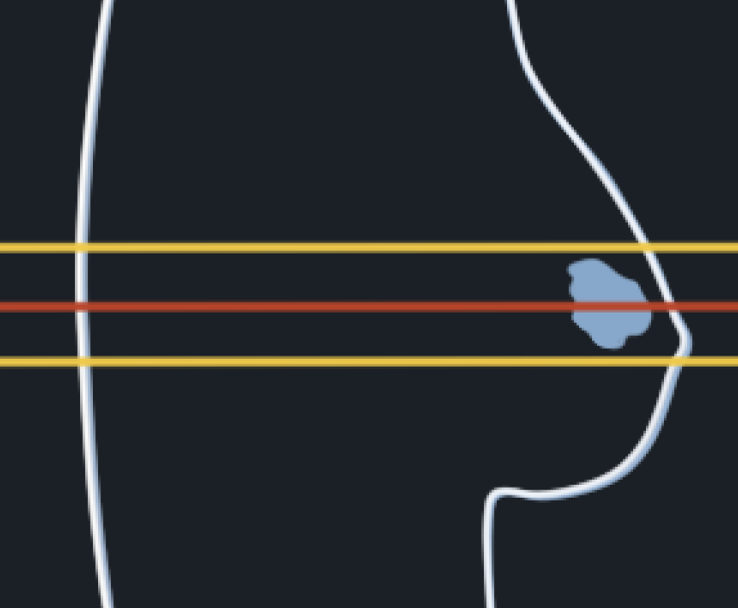}
\caption{Side profile}
\label{fig:MRI_chest_drawing2}
\end{subfigure}
\begin{subfigure}{0.715\textwidth}
\centering
\includegraphics[width=\textwidth]{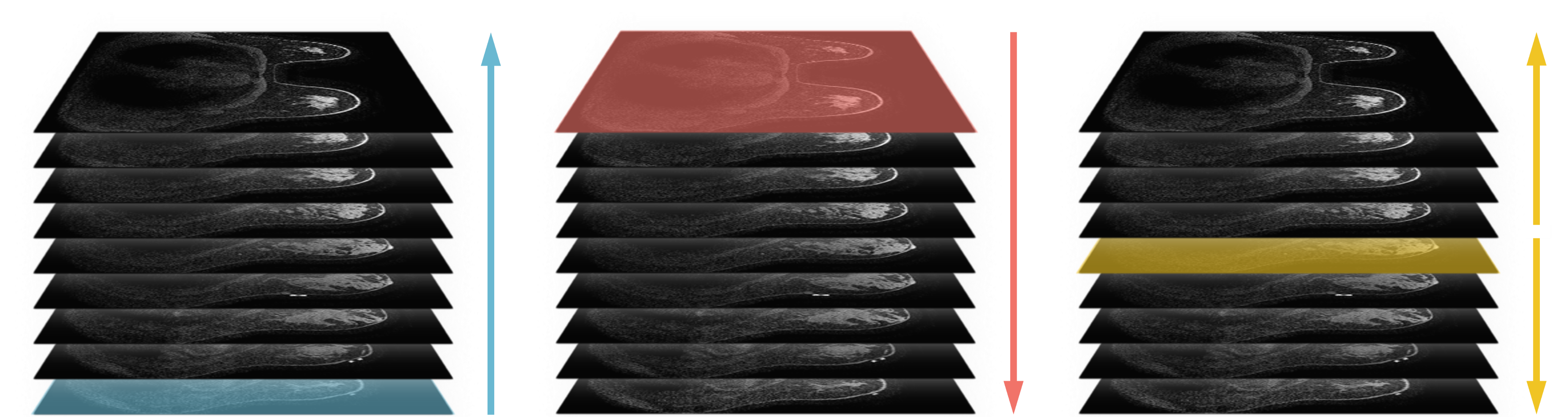}
\caption{Image propagation}
\label{fig:data_bot}
\end{subfigure}
\caption{(a) A side profile diagram of the breast, highlighting tumor-positive slices bounded by two yellow lines. The center slice of the tumor-positive region is depicted with a red line. (b) Visualization of propagation strategies employed in this work, each illustrated by the arrows from their respective starting slice.}
\label{fig:propagation_overview}
\end{figure}

\subsection{3D Tumor Segmentation}

To perform 3D tumor segmentation, we treat the volumetric breast MRI scan as a sequence of 2D axial slices. SAM2 is applied in a slice-wise fashion, where a segmentation mask is generated for each slice using the model's tracking functionality. The process begins with a single bounding box prompt provided on one slice, which serves as the initialization for the segmentation. From this initial input, SAM2 tracks the object across adjacent slices by using the predicted mask from the previous slice as contextual guidance for the next.

We evaluate three propagation strategies for traversing the 3D volume below and illustrate them in \figurename~\ref{fig:propagation_overview}.

\textbf{Bottom-to-top}. Starting from the bottom-most slice containing the tumor, we provide a bounding box prompt and sequentially propagate the segmentation upward. Each prediction uses the previous mask as a prompt.

\textbf{Top-to-bottom}. This approach mirrors the bottom-to-top strategy, but starts from the top-most tumor slice and moves downward through the volume.
        
\textbf{Center-outward}. Segmentation begins at the central slice of the tumor, which typically exhibits the largest and clearest tumor region. From this central point, propagation proceeds both upward and downward. This strategy aims to take advantage of the most reliable initial mask and reduce tracking errors over long ranges.

\textbf{Volumetric Dice Similarity Coefficient}. To evaluate the performance of our segmentation model, we use the Volumetric Dice Similarity Coefficient, a standard metric for measuring spatial overlap between the predicted segmentation and the ground truth in 3D medical imaging.

Given a predicted binary volume \(\mathtt{P} \in \{0, 1\}^{D \times W \times H}\) and a corresponding ground truth volume \(\mathtt{G} \in \{0, 1\}^{D \times W \times H}\), the volumetric Dice score is defined as:

\begin{equation}
\text{Dice}(\mathtt{P}, \mathtt{G}) = \frac{2 \cdot |\mathtt{P} \cap \mathtt{G}|}{|\mathtt{P}| + |\mathtt{G}|} \,,
\end{equation}

where \(|\mathtt{P} \cap \mathtt{G}|\) denotes the number of voxels where the prediction and ground truth both label the same voxel as foreground (i.e., true positives), and \(|\mathtt{P}|\) and \(|\mathtt{G}|\) are the total number of foreground voxels in the prediction and ground truth volumes, respectively.

This formulation aggregates all voxel-level predictions across the entire 3D volume before computing the score, providing a global measure of overlap. The Dice score ranges from 0 (no overlap) to 1 (perfect agreement) and is particularly robust in settings with imbalanced class distributions, such as tumor segmentation where the foreground occupies a small fraction of the scan.

\begin{figure}[t!]
\centering
\begin{subfigure}{0.59\textwidth}
\centering
\includegraphics[width=\textwidth]{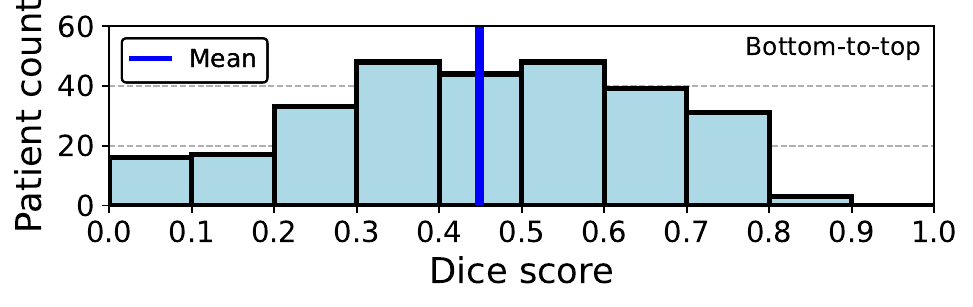}
\includegraphics[width=\textwidth]{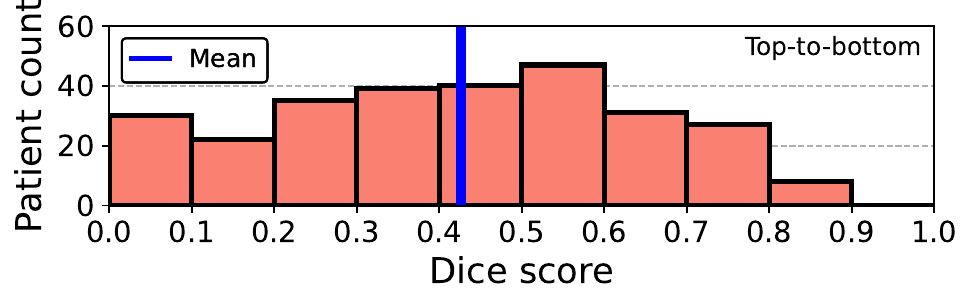}
\includegraphics[width=\textwidth]{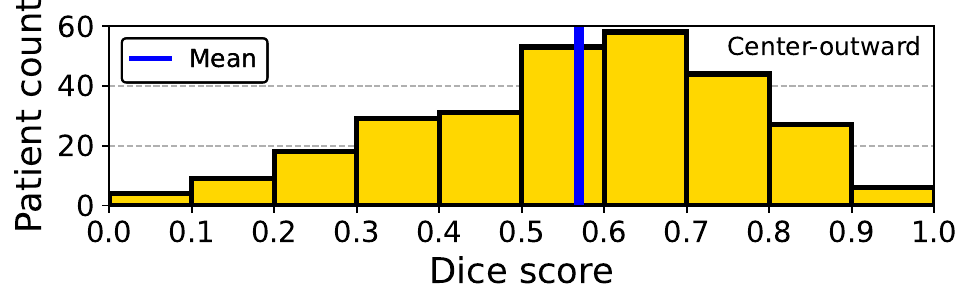}
\caption{}
\label{fig:exp_histograms}
\end{subfigure}
\begin{subfigure}{0.38\textwidth}
\centering
\begin{subfigure}{\textwidth}
\centering
\includegraphics[width=\textwidth]{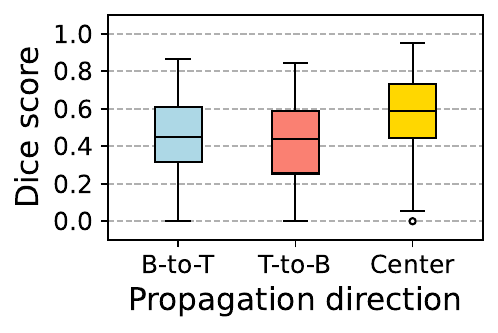}
\caption{}
\label{fig:exp_boxplots}
\end{subfigure}
\begin{subfigure}{\textwidth}
\centering
\includegraphics[width=\textwidth]{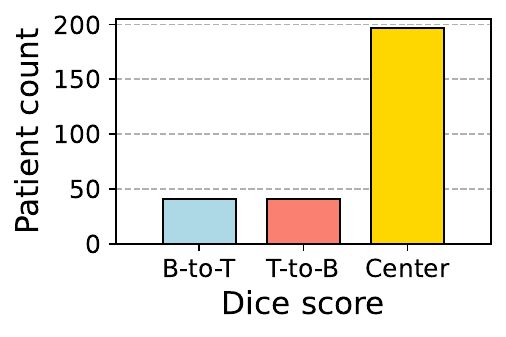}
\caption{}
\label{fig:exp_counts}
\end{subfigure}
\end{subfigure}
\caption{Evaluation of 3D propagation strategies. (a) Dice score histograms per propagation strategy with mean lines. (b) Boxplots of Dice scores grouped by propagation strategy. (c) Patient counts with highest Dice score per strategy.}
\label{fig:exp_results}
\end{figure}

\section{Experimental Results}
\label{sec:experiments}

\textbf{Quantitative Results}. 
We evaluate segmentation performance using the setup described in Section~\ref{sec:methodology}, where a single bounding box is used as input for tracking-based propagation across the 3D volume. As described, we employ three propagation strategies: bottom-to-top, top-to-bottom, and center-outward. The resulting Dice score distributions across patients are shown in \figurename~\ref{fig:exp_histograms}, with the mean score for each method indicated in blue lines. To compare the overall performance of the three strategies, we further aggregate these results and plot the Dice scores for all patients using boxplots in \figurename~\ref{fig:exp_boxplots}. Finally, we present a histogram showing the number of patients for which each propagation strategy achieved the highest Dice score in \figurename~\ref{fig:exp_counts}, providing insight into the consistency of each method across the dataset.

Among the three propagation strategies, the center-outward method outperforms the bottom-to-top and top-to-bottom approaches, showing higher median Dice scores in \figurename~\ref{fig:exp_boxplots} and a more favorable right-skewed distribution in \figurename~\ref{fig:exp_histograms}. This trend is further confirmed in \figurename~\ref{fig:exp_counts}, where the center-outward method yields the best performance for the majority of patients. These findings suggest that initializing segmentation from the central slice provides a more stable and effective propagation path. This is likely because the tumor tends to be largest and most clearly defined in the center, making it easier for the model to track its boundaries outward compared to starting from peripheral slices. Overall, the segmentation results are encouraging, especially considering that the SAM2 model was not explicitly trained for volumetric breast MRI data and that the only input provided is a single-slice bounding box mask. The ability to achieve robust 3D segmentations under such minimal supervision highlights the potential of leveraging foundation models such as SAM2 in data-scarce or low-resource clinical settings.

\begin{figure}[t!]
\centering
\begin{subfigure}{0.62\textwidth}
\centering
\includegraphics[width=\textwidth]{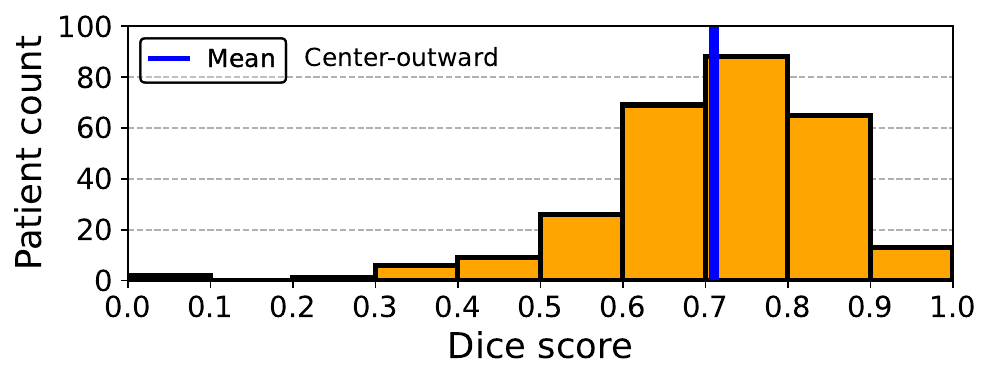}
\end{subfigure}
\begin{subfigure}{0.355\textwidth}
\centering
\includegraphics[width=\textwidth]{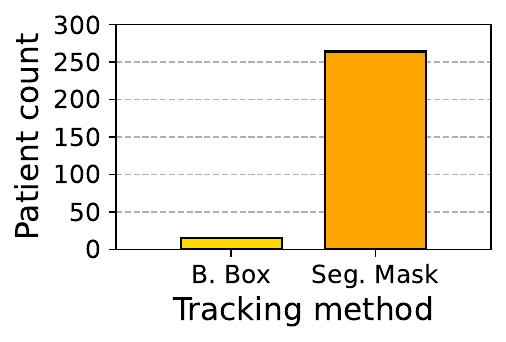}
\end{subfigure}
\caption{Evaluation of tumor tracking methods for SAM2. The histogram (left) shows the distribution of Dice scores across patients. The bar plot (right) compares the number of patients with higher Dice scores when using bounding boxes versus segmentation masks as prompts for SAM2 tracking.}
\label{fig:segmentation_mask_results}
\end{figure}

\begin{figure}[t!]
\centering
\begin{subfigure}{\textwidth}
\centering
\includegraphics[width=0.19\textwidth]{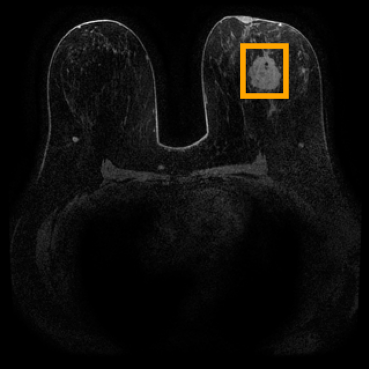}
\includegraphics[width=0.19\textwidth]{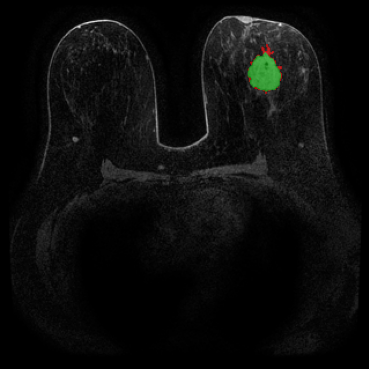}
\includegraphics[width=0.19\textwidth]{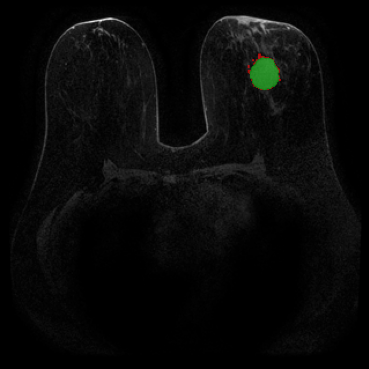}
\includegraphics[width=0.19\textwidth]{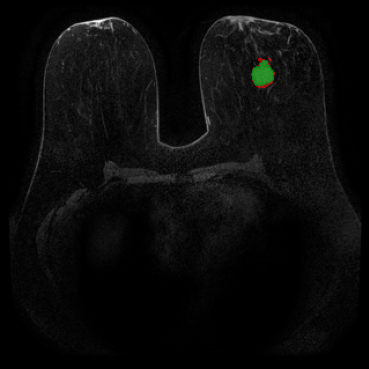}
\includegraphics[width=0.19\textwidth]{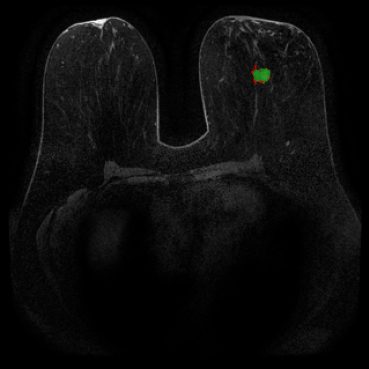}
\caption{Segmentation predictions for patient 350}
\label{fig:example1}
\end{subfigure}
\begin{subfigure}{\textwidth}
\centering
\includegraphics[width=0.19\textwidth]{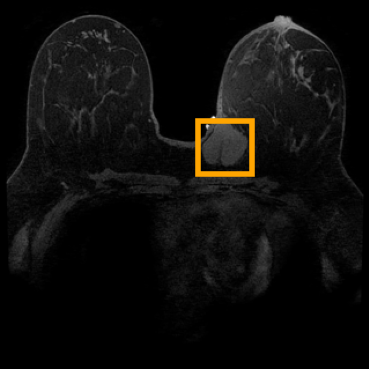}
\includegraphics[width=0.19\textwidth]{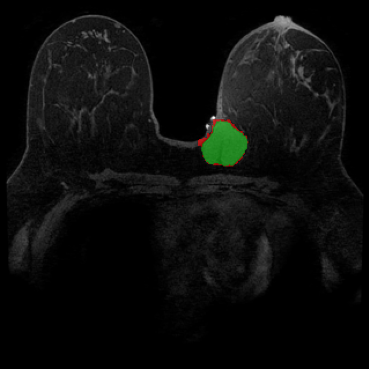}
\includegraphics[width=0.19\textwidth]{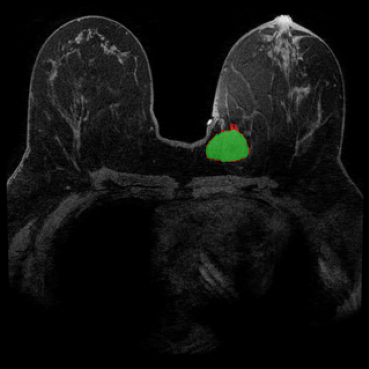}
\includegraphics[width=0.19\textwidth]{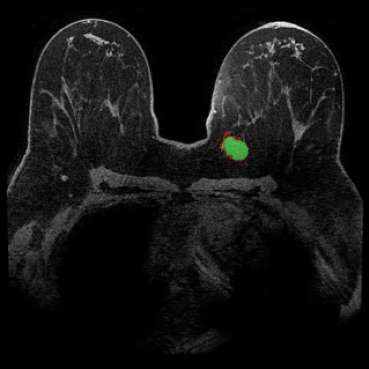}
\includegraphics[width=0.19\textwidth]{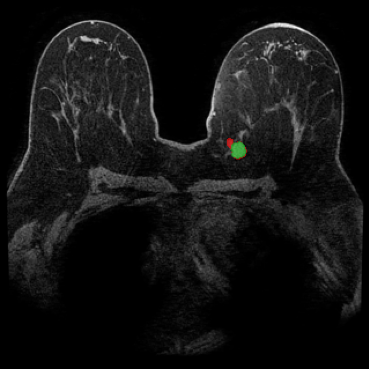}
\caption{Segmentation predictions for patient 400}
\label{fig:example2}
\end{subfigure}
\begin{subfigure}{0.45\textwidth}
\centering
\includegraphics[width=0.45\textwidth]{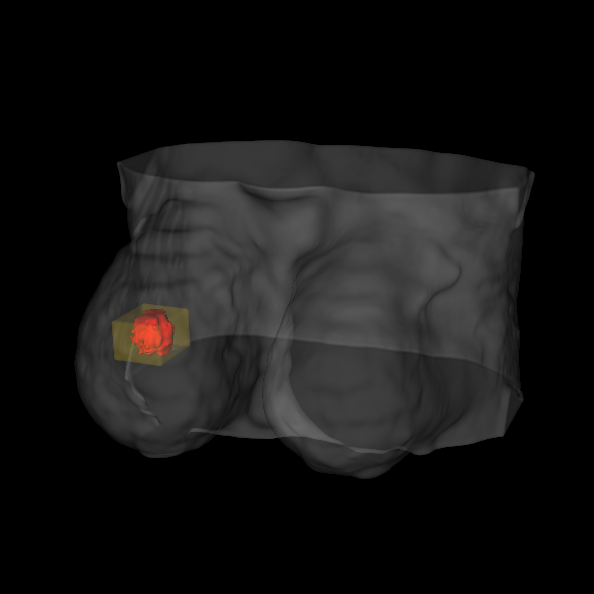}
\includegraphics[width=0.45\textwidth]{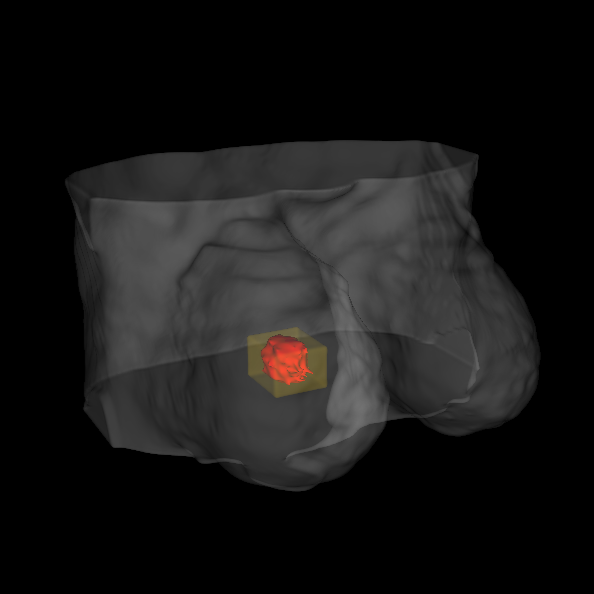}
\caption{3D view for patient 350}
\label{fig:example1_3D}
\end{subfigure}
\begin{subfigure}{0.45\textwidth}
\centering
\includegraphics[width=0.45\textwidth]{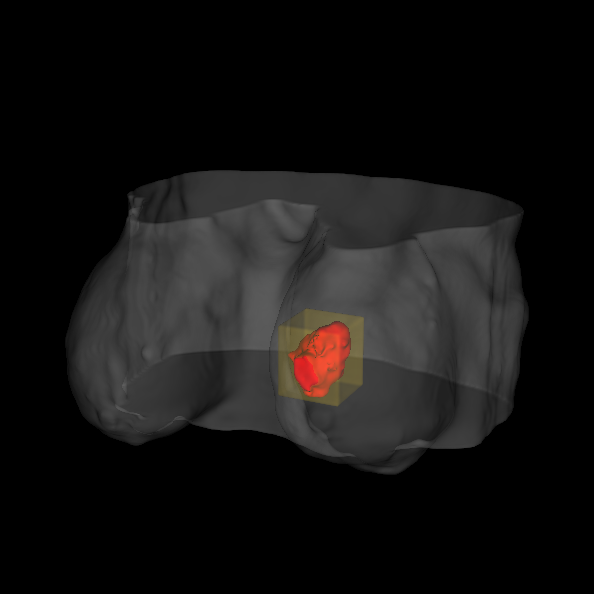}
\includegraphics[width=0.45\textwidth]{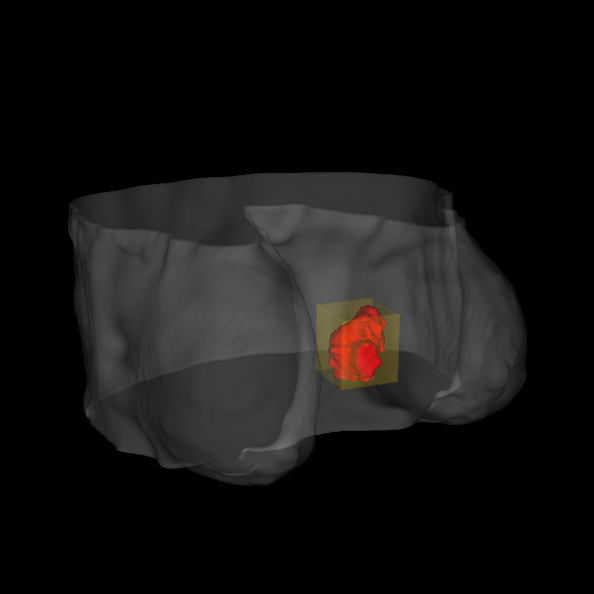}
\caption{3D view for patient 400}
\label{fig:example2_3D}
\end{subfigure}
\caption{Sequential MRI slices for (a) Patient 350 and (b) Patient 400. The leftmost slice shows the input with an orange bounding box indicating the tumor region. The next four slices show SAM2 tracking results (center-outward), where green denotes overlap between prediction and ground truth, and red indicates disagreement. (c) and (d) show 3D visualizations of the predicted tumor (red) within the input bounding box (orange).}
\label{fig:output_images}
\end{figure}

\textbf{Comparison to Mask-based Tracking}. 
To compare the performance of bounding box tracking with segmentation mask tracking, we rerun the center-outward propagation experiment by providing a segmentation mask on a single central slice as input, as opposed to bounding box. From this initial mask, the model tracks and propagates the segmentation to the remaining slices, using each predicted mask to guide the next step. These results are shown in \figurename~\ref{fig:segmentation_mask_results}.

For nearly all patients, segmentation accuracy improves with mask-based guidance compared to bounding box tracking, with only a few exceptions. This mask-based tracking provides a more precise and context-aware prior than the coarse bounding box approach. As a result, we observe a substantial improvement in segmentation quality, with the mean Dice score across all patients increasing from 0.57 (bounding box tracking) to 0.71 (segmentation mask tracking). This improvement highlights the benefit of using more informative guidance during propagation, particularly in complex anatomical regions where bounding boxes may fail to represent tumor boundaries accurately, further validating the potential of SAM2-based workflows under minimal supervision.

\textbf{Qualitative Results}. 
To illustrate the segmentation performance qualitatively, we present example predictions in \figurename~\ref{fig:output_images}, including selected 2D slices and 3D reconstructions that highlight the predicted tumor regions. These visualizations demonstrate that the model is able to produce coherent and anatomically plausible segmentations across slices. Additional 3D visualizations, which further support the consistency and quality of the predictions, are included in the supplementary material.

\begin{figure}[t!]
\centering
\begin{subfigure}{0.99\textwidth}
\centering
\includegraphics[width=0.315\textwidth]{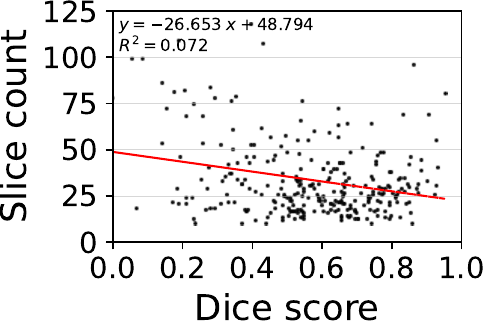}
\includegraphics[width=0.33\textwidth]{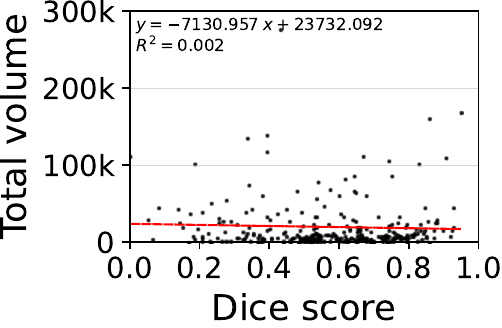}
\includegraphics[width=0.315\textwidth]{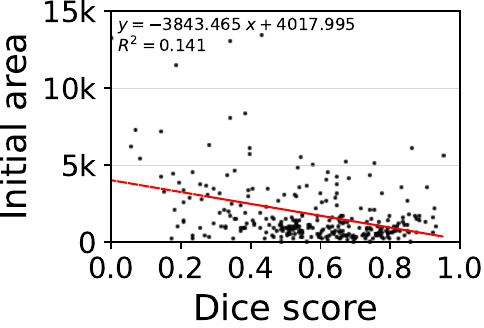}
\end{subfigure}
\caption{Scatter plots showing relationships between Dice score and tumor properties. Each plot shows Dice score on the x-axis and a tumor-related variable on the y-axis: (left) number of slices containing tumor, (middle) total tumor volume in voxels, (right) initial tumor area in the tracking slice. A linear regression line is fitted in red, with corresponding equation and \( R^2 \) value displayed in each subplot.}
\label{fig:scatter_plots}
\end{figure}

\textbf{Factors Affecting Dice Score}. 
To explore factors associated with successful and unsuccessful segmentations, we analyze the correlation between the volumetric Dice score and three variables: (1) the number of slices containing tumor, (2) the total tumor volume, and (3) the initial tumor area in the tracking frame. The results are shown in \figurename~\ref{fig:scatter_plots}. As the fitted trend lines and low \( R^2 \) values indicate, none of these variables show a meaningful correlation with segmentation performance.

\begin{figure}[t!]
\centering
\includegraphics[width=0.19\textwidth]{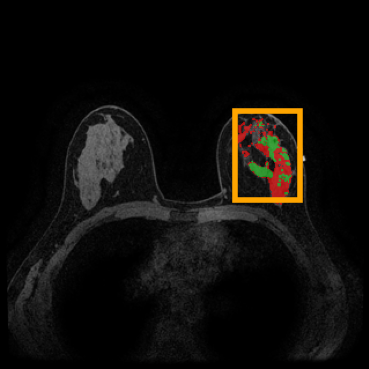}
\includegraphics[width=0.19\textwidth]{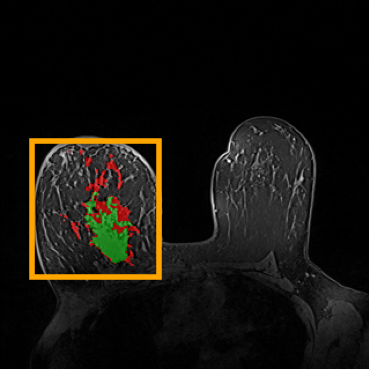}
\includegraphics[width=0.19\textwidth]{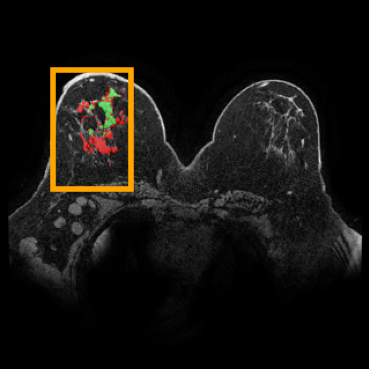}
\includegraphics[width=0.19\textwidth]{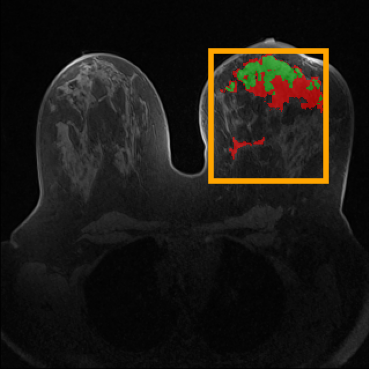}
\includegraphics[width=0.19\textwidth]{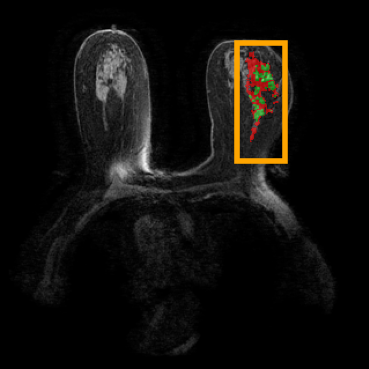}
\caption{Example cases where SAM2 yields low Dice scores due to the presence of multiple small lesions, rather than a single large, well-defined tumor.}
\label{fig:bad_predictions}
\end{figure}

\section{Conclusion and Future Perspectives}

In this study, we demonstrate the effective adaptation of SAM2 for invasive MRI breast tumor segmentation with minimal human input. Evaluating three propagation strategies, we found that center-outward initialization results in more accurate and stable segmentation output compared to the other two approaches. These results, achieved with only a single bounding box annotation per patient, highlight the potential of vision foundation models to perform robust volumetric medical imaging segmentation, which they were not originally designed or trained for.

As discussed in Section~\ref{sec:experiments}, we investigated a number of properties of MRIs to discover cases where SAM2 fails to segment tumors accurately. Our analysis did not reveal any strong correlation between segmentation performance and factors such as tumor volume, number of affected slices, or initial tumor area. However, upon conducting a qualitative analysis, we discovered that segmentation failures commonly occurred in cases with multiple small lesions scattered across slices, rather than a single large, well-defined tumor (see \figurename~\ref{fig:bad_predictions}). These cases challenge the model's ability to maintain spatial coherence during propagation. In future work, we believe that incorporating lightweight pre-selection strategies to better handle fragmented tumor presentations, or integrating uncertainty-aware mechanisms, could further improve segmentation robustness under minimal supervision.

Our work opens up several promising directions for future research. One natural extension is to evaluate and compare the performance of other segmentation foundation models, such as MedSAM2~\cite{ma2025medsam2}, which are specifically adapted for medical images and volumetric data. Furthermore, combining SAM2 with lightweight pre-processing modules for lesion detection or integrating temporal consistency mechanisms may enhance segmentation robustness in complex cases. Exploring domain adaptation strategies or fine-tuning with a small amount of annotated data could also improve performance while maintaining accessibility in low-resource settings.

\bibliographystyle{splncs04}
\bibliography{DeepBreath_42}
\end{document}